\documentclass[runningheads]{llncs}
\pdfoutput=1
 
\usepackage{eccv}

\usepackage{eccvabbrv}

\usepackage{graphicx}
\usepackage{booktabs}
\usepackage{threeparttable}

\usepackage[accsupp]{axessibility}  

\usepackage{hyperref}

\usepackage{orcidlink}

\usepackage[textsize=scriptsize]{todonotes}

\newenvironment{icompact}{
  \begin{list}{$\bullet$}{
    \itemindent -.05em
    \parsep 0pt plus 1pt
    \partopsep 0pt plus 1pt
    \topsep 2pt plus 2pt minus 2pt
    \itemsep 0pt plus 1.3pt
    \parskip 0pt plus 2pt
    \leftmargin 0.13in}
      }
{\normalsize
\end{list}
}

\usepackage{amssymb}
\usepackage{amsmath,amsfonts}
\usepackage{graphicx}
\usepackage{subcaption}
\usepackage{caption}
\usepackage{textcomp}
\usepackage{verbatim}
\usepackage{listings}
\usepackage{xcolor}

\usepackage{xcolor}
\hypersetup{
    colorlinks=true,
    linkcolor=blue,
    filecolor=magenta,      
    urlcolor=cyan,
    pdftitle={Overleaf Example},
    pdfpagemode=FullScreen,
    }

\newcommand{\sys}{{AnomalyRuler}\xspace}
\newcommand{\sysbase}{{AnomalyRuler-base}\xspace}

\usepackage{amssymb}
\usepackage{pifont}
\newcommand{\cmark}{\ding{51}}%
\newcommand{\xmark}{\ding{55}}%
\usepackage{wrapfig}
\usepackage{multirow}
\usepackage{colortbl} 
\newcommand{\highlight}[1]{{\setlength{\fboxsep}{0pt}\colorbox{yellow!50}{#1}}} 

\usepackage{microtype}
\usepackage{comment}

\usepackage{bbm}

\begin{document}

\title{\Large Follow the Rules: Reasoning for Video Anomaly Detection with Large Language Models} 

\titlerunning{Follow the Rules: Reasoning for VAD with LLMs}

\author{Yuchen Yang \inst{1}\thanks{This work was mostly done when Y. Yang was an intern at HRI-USA.} \and
Kwonjoon Lee \inst{2} \and
Behzad Dariush \inst{2} \and
Yinzhi Cao \inst{1} \and \\
Shao-Yuan Lo \inst{2}}

\authorrunning{Y. Yang et al.}

\institute{Johns Hopkins University \\
\email{\{yc.yang, yinzhi.cao\}@jhu.edu} \and
Honda Research Institute USA \\
\email{\{kwonjoon\_lee, bdariush, shao-yuan\_lo\}@honda-ri.com}}

\maketitle

\begin{abstract}

Video Anomaly Detection (VAD) is crucial for applications such as security surveillance and autonomous driving. However, existing VAD methods provide little rationale behind detection, hindering public trust in real-world deployments. In this paper, we approach VAD with a reasoning framework. Although Large Language Models (LLMs) have shown revolutionary reasoning ability, we find that their direct use falls short of VAD. Specifically, the implicit knowledge pre-trained in LLMs focuses on general context and thus may not apply to every specific real-world VAD scenario, leading to inflexibility and inaccuracy. To address this, we propose \sys, a novel rule-based reasoning framework for VAD with LLMs. \sys comprises two main stages: induction and deduction. In the induction stage, the LLM is fed with few-shot normal reference samples and then summarizes these normal patterns to induce a set of rules for detecting anomalies. The deduction stage follows the induced rules to spot anomalous frames in test videos. Additionally, we design rule aggregation, perception smoothing, and robust reasoning strategies to further enhance \sys's robustness. \sys is the first reasoning approach for the one-class VAD task, which requires only few-normal-shot prompting without the need for full-shot training, thereby enabling fast adaption to various VAD scenarios. Comprehensive experiments across four VAD benchmarks demonstrate \sys's state-of-the-art detection performance and reasoning ability. \sys is open-source and available at: \url{https://github.com/Yuchen413/AnomalyRuler}

\vspace{-0.1in}

\end{abstract}

\begin{figure}[t!]
\centering
\includegraphics[width=0.98\linewidth]{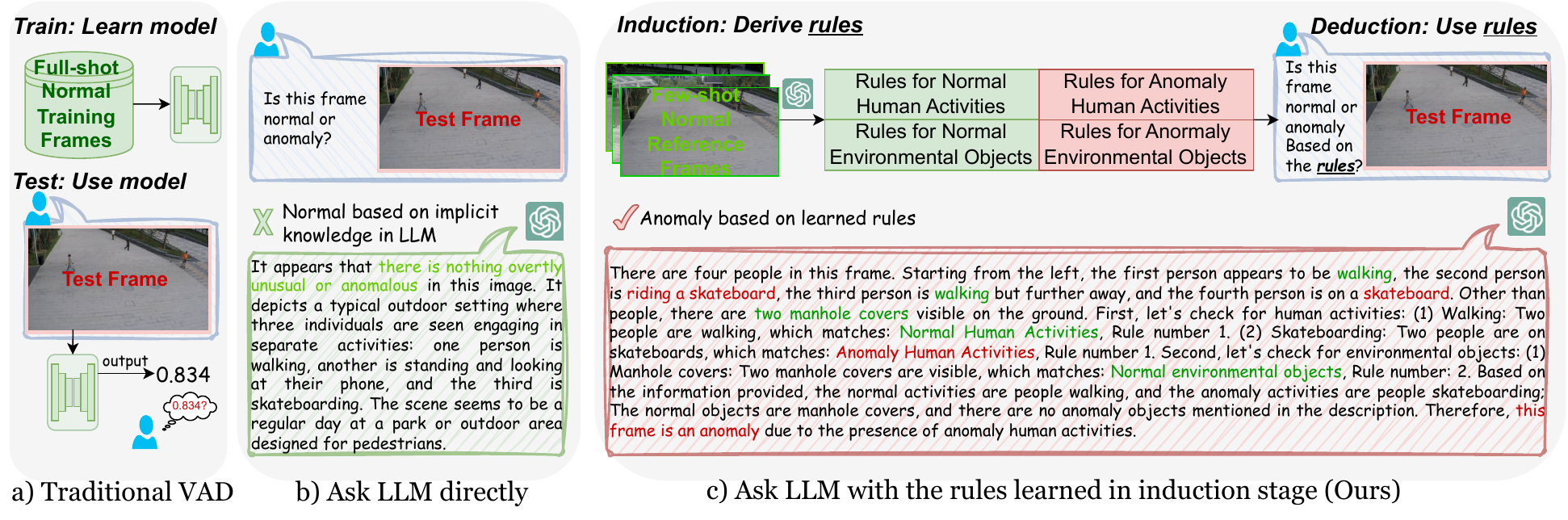}
\caption{Comparison of one-class VAD approaches. In this specific safety application example, only ``walking'' is normal. The test frame contains ``skateboarding'', so it is abnormal. (a) Traditional methods require full-shot training and only output anomaly scores, lacking reasoning. (b) Direct LLM use may not align with specific VAD needs. Here GPT-4V mistakenly treats ``skateboarding'' as normal. (c) Our \sys has induction and deduction stages. It derives rules from few-shot normal reference frames to detect anomalies, correctly identifying ``skateboarding'' as an anomaly.}\label{fig:intro}
\end{figure}

\section{Introduction}

Video Anomaly Detection (VAD) aims to identify anomalous activities, which are infrequent or unexpected in surveillance videos. It has a wide range of practical applications, including security (e.g., violence), autonomous driving (e.g., traffic accidents), etc. VAD is a challenging problem since anomalies are rare and long-tailed in real life, leading to a lack of large-scale representative anomaly data. Hence, the one-class VAD (a.k.a. unsupervised VAD) paradigm ~\cite{park2020learning_memg,VADRE_shi2023video,VADRE_sun2023_scene,vadre_hasan2016learning,wu2022self} is preferred, as it assumes that only the more accessible normal data are available for training. Most existing one-class VAD methods learn to model normal patterns via self-supervised pretext tasks, such as frame reconstruction~\cite{VADRE_liu2021hf2vad,lo2022adversarially,park2020learning_memg,VADRE_sun2023_scene,vadre_hasan2016learning,wu2022self, VADRE_Yan_2023_ICCV} and frame order classification~\cite{VADDI_georgescu2021background,VADRE_shi2023video,VADRE_wang2022video}. Despite good performance, these traditional methods can only output anomaly scores, providing little rationale behind their detection results (see Fig.~\ref{fig:intro}a). This hinders them from earning public trust when deployed in real-world products.

We approach the VAD task with a reasoning framework toward a trustworthy system, which is less explored in the literature. An intuitive way is to incorporate the emergent Large Language Models (LLMs)~\cite{llm_achiam2023gpt4,brown2020language,llm_jiang2023mistral,radford2018improving,radford2019language,llm_touvron2023llama2}, which have shown revolutionary capability in various reasoning tasks. Still, we find that their direct use falls short of performing VAD. Specifically, the implicit knowledge pre-trained in LLMs focuses on general context, meaning that it may not always align with specific real-world VAD applications. In other words, there is a mismatch between an LLM’s understanding of anomalies and the anomaly definitions required for certain scenarios. For example, the GPT-4V~\cite{llm_achiam2023gpt4} typically treats ``skateboarding'' as a normal activity, whereas certain safety applications need to define it as an anomaly, such as within a restricted campus (see Fig.~\ref{fig:intro}b). However, injecting such specific knowledge by fine-tuning LLMs for each application is costly. This highlights the necessity for a flexible prompting approach that steers LLMs' reasoning strengths to different uses of VAD.

To arrive at such a solution, we revisit the fundamental process of the scientific method~\cite{bacon1620} emphasizing reasoning, which involves drawing conclusions in a rigorous manner~\cite{seel2011}. Our motivation stems from two types of reasoning: inductive reasoning, which infers generic principles from given observations, and deductive reasoning, which derives conclusions based on given premises. In this paper, we propose \sys, a new VAD framework based on reasoning with LLMs. \sys consists of an induction stage and a deduction stage as shown in Fig.~\ref{fig:intro}c. In the induction stage, the LLM is fed with visual descriptions of few-shot normal samples as references to derive a set of rules for determining normality. Here we employ a Vision-Language Model (VLM)~\cite{lvm_liu2023llava,lvm_wang2023cogvlm} to generate the description for each input video frame. Next, the LLM derives a set of rules for detecting anomalies by contrasting the rules for normality. The deduction, which is also an inference stage, follows the induced rules to identify anomalous frames in test video sequences. Additionally, in response to potential perception and reasoning errors by the VLM and LLM, we design strategies including rule aggregation via the randomized smoothing~\cite{pmlr-v97-cohen19c-rr} for rule induction error mitigation, perception smoothing via the proposed Exponential Majority Smoothing for perception error reduction together with temporal consistency enhancement, and robust reasoning via a recheck mechanism for reliable reasoning output. These strategies are integrated into the \sys pipeline to further enhance its detection robustness.

Apart from equipping VAD with reasoning ability, \sys offers several advantages. First, \sys is a novel \emph{few-normal-shot prompting} approach that utilizes only a few normal samples from a training set as references to derive the rules for VAD. This avoids the need for expensive full-shot training or fine-tuning of the entire training set, as required by traditional one-class VAD methods. Importantly, it enables efficient adaption by redirecting LLM's implicit knowledge to different specific VAD applications through just a few normal reference samples. Second, \sys shows strong domain adaptability across datasets, as the language provides consistent descriptions across different visual domains, e.g., ``walking'' over visual data variance. This allows the application of induced rules to datasets with similar scenarios but distinct visual appearances. Furthermore, \sys is a generic framework that is complementary to VLM and LLM backbones. It accommodates both closed-source models such as the GPT family~\cite{llm_achiam2023gpt4,radford2019language} and open-source alternatives such as Mistral~\cite{llm_jiang2023mistral}. To the best of our knowledge, the proposed \sys is the first reasoning approach for the one-class VAD problem. Extensive experiments on four VAD datasets demonstrate \sys's state-of-the-art performance, reasoning ability, and domain adaptability.

In summary, this paper has three main contributions. (1) We propose a novel rule-based reasoning framework for VAD with LLMs, namely \sys. To the best of our knowledge, it is the first reasoning approach for one-class VAD. (2) The proposed \sys is a novel few-normal-shot prompting approach that eliminates the need for expensive full-shot tuning and enables fast adaption to various VAD scenarios. (3) We propose rule aggregation, perception smoothing, and robust reasoning strategies for \sys to enhance its robustness, leading to state-of-the-art detection performance, reasoning ability, and domain adaptability.

\section{Related Work}

\paragraph{Video Anomaly Detection.}
VAD is a challenging task since anomaly data are scarce and long-tailed. Therefore, researchers often focus on the one-class VAD (a.k.a. unsupervised VAD) paradigm~\cite{VADDI_georgescu2021background,VADRE_liu2021hf2vad,hirschorn2023normalizing,lo2022adversarially,park2020learning_memg,VADRE_shi2023video,VADRE_sun2023_scene,vadre_hasan2016learning,VADRE_wang2022video,wu2022self,VADRE_Yan_2023_ICCV}, which uses only normal data during training. Most one-class methods learn to model normal patterns via self-supervised pretext tasks, based on the assumption that the model would obtain poor pretext task performance on anomaly data. Reconstruction-based methods~\cite{VADRE_liu2021hf2vad,lo2022adversarially,park2020learning_memg,VADRE_sun2023_scene,vadre_hasan2016learning,wu2022self,VADRE_Yan_2023_ICCV} employ generative models such as auto-encoders and diffusion models to perform frame reconstruction or frame prediction as pretext tasks. Distance-based~\cite{VADDI_georgescu2021background,VADRE_shi2023video,VADRE_wang2022video} methods use classifiers to perform pretext tasks such as frame order classification. These traditional methods can only output anomaly scores, providing little rationale behind their detection. 
Several recent studies explore utilizing VLMs or LLMs in anomaly detection. Elhafsi et al.~\cite{elhafsi23_sad} analyze semantic anomalies with an object detector~\cite{minderer2022simple} an LLM~\cite{brown2020language} in driving scenes. However, it relies on predefined concepts of normality and anomaly, which limits its adaption to different scenarios and cannot handle long-tailed undefined anomalies. Moreover, this method has not been evaluated on standard VAD benchmarks~\cite{acsintoae2022ubnormal,li2013anomaly_ped2,liu2018future_sht,lu2013abnormal_ave}. Cao et al.~\cite{cao2023towards} explore the use of GPT-4V for anomaly detection, but their direct use may fall into the misalignment between GPT-4V’s implicit knowledge and specific VAD needs, as discussed. Gu et al.~\cite{gu2023anomalyagpt} adopt a large VLM for anomaly detection, but it focuses on industrial images. Despite supporting dialogues, this method can only describe anomalies rather than explain the rationales behind its detection. Lv et al.~\cite{lv2024video} equip video-based LLMs in the VAD framework to provide detection explanations. It involves three-phase training to fine-tune the heavy video-based LLMs. Besides, it focuses on weakly-supervised VAD, a relaxed paradigm that requires training with anomaly data and labels. Different from these works, our \sys provides rule-based reasoning via efficient few-normal-shot prompting and enables fast adaption to different VAD scenarios.

\paragraph{Large Language Models.}
LLMs~\cite{llm_achiam2023gpt4,brown2020language,llm_jiang2023mistral,radford2018improving,radford2019language,touvron2023llama,llm_touvron2023llama2} have achieved significant success in natural language processing and are recently being explored for computer vision problems. Recent advances, such as the GPT family~\cite{llm_achiam2023gpt4,brown2020language,radford2018improving,radford2019language}, the LLaMA family~\cite{touvron2023llama,llm_touvron2023llama2}, and Mistral~\cite{llm_jiang2023mistral}, have shown remarkable capabilities in understanding and generating human language. On the other hand, large VLMs~\cite{llm_achiam2023gpt4,lvm_li2023blip2,lvm_liu2023llava,mittal2024can,lvm_wang2023cogvlm,su2023pandagpt,zhang2023recognize,lvm_zhang2023videollama,lvm_zhu2023minigpt} have shown promise in bridging the vision and language domains. BLIP-2~\cite{lvm_li2023blip2} leverages Q-Former to integrate visual features into a language model. LLaVA~\cite{lvm_liu2023llava} introduces a visual instruction tuning method for visual and language understanding. CogVLM~\cite{lvm_wang2023cogvlm} trains a visual expert module to improve large VLM’s vision ability. Video-LLaMA~\cite{lvm_zhang2023videollama} extends LLMs to understand video data. These models’ parametric knowledge is trained for general purposes and thus may not apply to every VAD application. Recent studies explore prompting methods to exploit LLMs’ reasoning ability. Chain-of-Thought (CoT)~\cite{reasoning_wei2022cot,reasoning_diao2023activecot} guides LLMs to solve complex problems via multiple smaller and manageable intermediate steps. Least-to-Most (LtM)~\cite{zhou2023least,reasoning_devide} decomposes a complex problem into multiple simpler sub-problems and solves them in sequence. Hypotheses-to-Theories (HtT)~\cite{reasoning_zhu2023htt} learns a rule library for reasoning from labeled training data in a supervised manner. However, a reasoning approach for the VAD task in the one-class paradigm is not well-explored.

\section{Induction}\label{sec:in}
\begin{figure}[t!]
  \centering
  \includegraphics[width=0.98\linewidth]{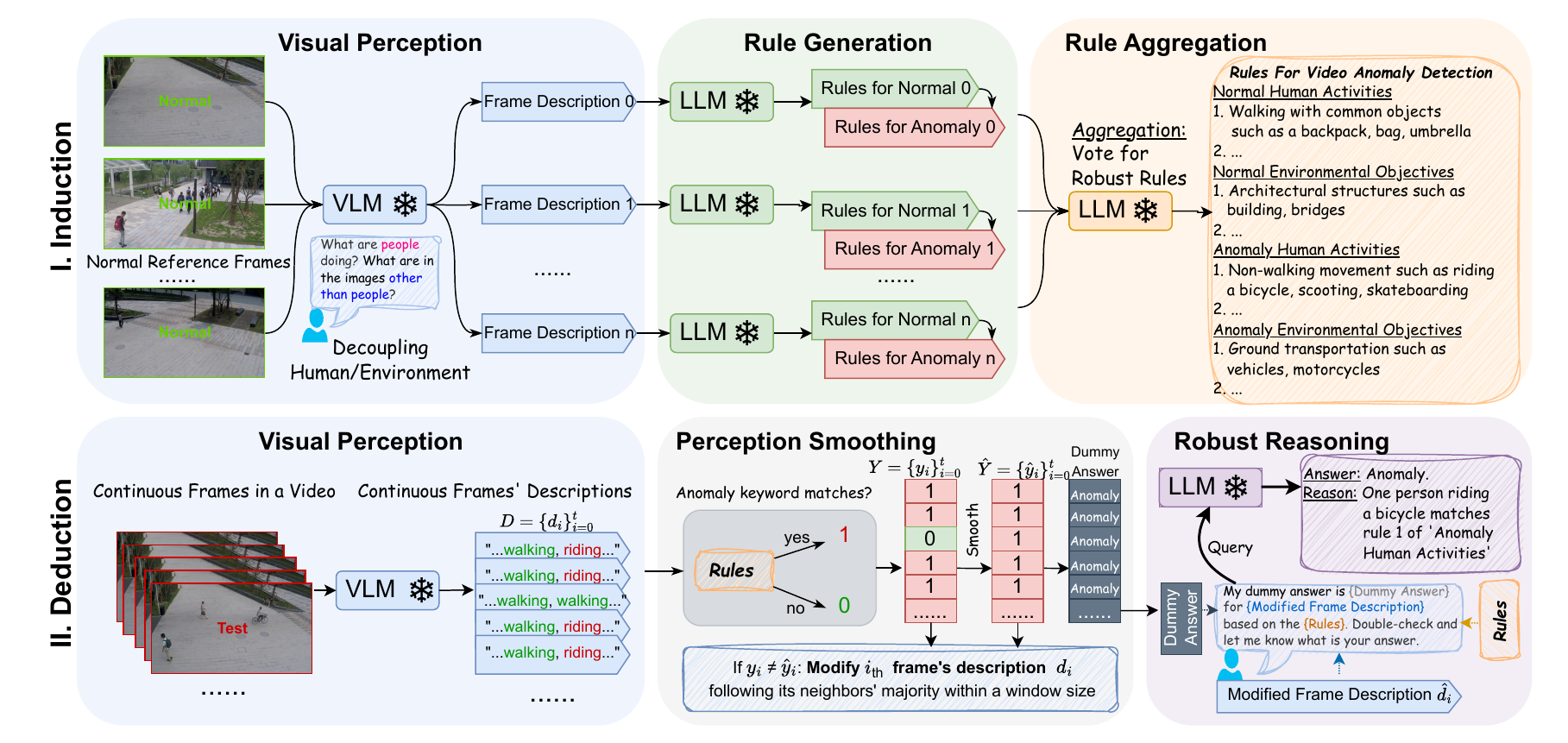} 
  \caption{The \sys pipeline consists of two main stages: induction and deduction. The induction stage involves: i) visual perception transfers normal reference frames to text descriptions; ii) rule generation derives rules based on these descriptions to determine normality and anomaly; iii) rule aggregation employs a voting mechanism to mitigate errors in rules. The deduction stage involves: i) visual perception transfers continuous frames to descriptions; ii) perception smoothing adjusts these descriptions considering temporal consistency to ensure neighboring frames share similar characteristics; iii) robust reasoning rechecks the previous dummy answers and outputs reasoning.} \label{fig:pipe}
\end{figure}

The induction stage aims to derive a set of rules from a few normal reference frames for performing VAD. The top part of Fig.~\ref{fig:pipe} shows the three modules in the induction pipeline. The visual perception module utilizes a VLM which takes a few normal reference frames as inputs and outputs frame descriptions. The rule generation module uses an LLM to generate rules based on these descriptions. The rule aggregation module employs a voting mechanism to mitigate the errors from rule generation. In the following sections, we discuss each module and the strategies applied in detail.

\subsection{Visual Perception}

We design the visual perception module as the initial step in our pipeline. This module utilizes a VLM to convert video frames into text descriptions. We define $F_{\text{normal}} = \{f_{\text{normal}_0}, \text{ } \ldots, \text{ } f_{\text{normal}_n}\}$ as the few-normal-shot reference frames, with each frame $f_{\text{normal}_i} \in F_{\text{normal}}$ randomly chosen from the training set. This module outputs the text description of each normal reference frame: 
$D_{\text{normal}} = \{\text{VLM}(f_{\text{normal}_i}, p_v) \mid f_{\text{normal}_i} \in F_{\text{normal}}\}$, with $p_v$ as the prompt ``What are people doing? What are in the images other than people?''. Instead of directly asking ``What are in the image?'', we design $p_v$ to separate humans and the environment with the following advantages. First, it enhances perception precision by directing the model's attention to specific aspects of the scene, ensuring that no details are overlooked. Second, it simplifies the following rule generation module by dividing the task into two subproblems\cite{reasoning_devide}, i.e., rules for human activities and rules for environmental objects. We denote this strategy as Human and Environment.

\subsection{Rule Generation}

With the text descriptions from normal reference frames $D_{\text{normal}}$, we design a Rule Generation module that uses a frozen LLM to generate rules (denoted as $R$). In formal terms, $R = \{\text{LLM}(d_{\text{normal}_i}, p_{g}) \mid d_{\text{normal}_i} \in D_{\text{normal}}\}$, where $p_g$ is the prompt detailed in Appendix~\ref{appendix:prompts}. We craft $p_g$ with three strategies to guide the LLM in gradually deriving rules from the observed normal patterns:

\paragraph{Normal and Anomaly.} 
The prompt $p_g$ guides the LLM to perform \emph{contrast}, which first induces rules for normal based on $D_{\text{normal}}$, which are assumed to be ground-truth normal. Then, it generates rules for anomalies by contrasting them with the rules for normal. For instance, if ``walking'' is a common pattern in $D_{\text{normal}}$, it becomes a normal rule, and then ``non-walking movement'' will be included in the rules for anomaly. This strategy sets a clear boundary between normal and anomaly without access to anomaly frames.
\paragraph{Abstract and Concrete.}
The prompt $p_g$ helps the LLM to perform \emph{analogy}, which starts from an abstract concept and then effectively generalizes to more concrete examples. Taking the same ``walking'' example, the definition of a normal rule is now expanded to ``walking, whether alone or with others.'' Consequently, the anomaly rule evolves to include specific non-walking movements, i.e., ``non-walking movement, such as riding a bicycle, scooting, or skateboarding.'' This strategy clarifies the rules with detailed examples and enables the LLM to use analogy for reasoning without exhaustively covering every potential scenario.
\paragraph{Human and Environment.} This strategy is inherited from the Visual Perception module. The prompt $p_g$ leads the LLM to pay attention separately to environmental elements (e.g., vehicles or scene factors) and human activities, separately. This enriches the rule set for VAD tasks, where anomalies often arise from interactions between humans and their environment.

\vspace{0.05in}
\noindent{These strategies align with the spirit of CoT~\cite{reasoning_wei2022cot} yet are further refined for the VAD task. The ablation study in Section~\ref{sec:ab} demonstrates their effectiveness.}

\subsection{Rule Aggregation}\label{sec:rule_aggregtion}

The rule aggregation module uses a LLM as an aggregator with a voting mechanism to combine $n$ sets of rules (i.e., $R$) generated independently from $n$ randomly chosen normal reference frames into one set of robust rules, $R_{\text{robust}} = \text{LLM}(R, p_{a})$. This module aims to mitigate errors from previous stages, such as the visual perception module's potential misinterpretation of ``walking'' as ``skateboarding'', leading to incorrect rules. The aggregation process filters out uncommon elements by retaining rule elements consistently present across the $n$ sets. The prompt $p_a$ for the LLM to achieve this is detailed in Appendix~\ref{appendix:prompts}. This strategy is based on the assumption of ~\emph{randomize smoothing}~\cite{pmlr-v97-cohen19c-rr}, where errors may occur on a single input but are less likely to consistently occur across multiple randomly sampled inputs. Therefore, by aggregating these outputs, \sys generates rules more resilient to individual errors. The hyperparameter $n$ can be treated as the number of batches. For simplicity, previous discussions assume that each batch has only one frame, i.e., $m=1$. Here we define $m$ as the number of normal reference frames per batch, i.e., batch size. We show the effectiveness of the rule aggregation and provide an ablation on different $n$ and $m$ values in Section~\ref{sec:ab}.

\section{Deduction}\label{sec:de}

After the induction stage derives a set of robust rules, the deduction stage follows these rules to perform VAD. The bottom part of Fig.~\ref{fig:pipe} illustrates the deduction stage, which aims to precisely perceive each frame of videos and then use the LLM to reason if they are normal or abnormal based on the rules. To achieve this goal, we design three modules. First, the visual perception module works similarly as described in the induction stage. However, instead of taking the few-normal-shot reference frames, the deduction processes continuous frames from each test video and outputs a series of frame descriptions $D = \{d_0, d_1, \ldots, d_t\}$. Second, the perception smoothing module reduces errors with the proposed Exponential Majority Smoothing. This step alone can provide preliminary detection results, referred to as \sysbase. Third, the robust reasoning module utilizes an LLM to recheck the preliminary detection results against the rules and perform reasoning. The perception smoothing and robust reasoning modules are elaborated in the following sections.

\subsection{Perception Smoothing}

As we discussed in Section~\ref{sec:rule_aggregtion}, visual perception errors would happen in the induction stage, and this concern extends to the deduction stage as well. To address this challenge, we propose a novel mechanism named Exponential Majority Smoothing. This mechanism mitigates the errors by considering temporal consistency in videos, i.e., movements are continuous and should exhibit consistent patterns over time. We utilize the results of this smoothing to guide the correction of frame descriptions, enhancing \sys's robustness to errors. There are four key steps:

\paragraph{Initial Anomaly Matching.}
For the continuous frame descriptions $D = \{d_0, d_1, \ldots, d_t\}$, \sys first match anomaly keywords $K$ found within the anomaly rules from the induction stage (see details in Appendix~\ref{appendix:prompts}), and assigns $d_i$ with label $y_i$ where $i \in [0,t]$, represents the predicted label. Formally, we have $y_i = 1$ if $\exists k \in K \subseteq d_i$, indicating an anomaly triggered by keywords such as ing-verb ``riding'' or ``running''. Otherwise, $y_i = 0$ indicates the normal. We denote the initial matching predictions as $Y = \{y_0, y_1, \ldots, y_t\}$.

\paragraph{Exponential Majority Smoothing.} 
We propose an approach that combines Exponential Moving Average (EMA) and Majority Vote. This approach is designed to enhance the continuity in human or object movements by adjusting the predictions to reflect the most common state within a specified window. The final smoothed predictions are denoted as $\hat{Y} = \{\hat{y}_0, \hat{y}_1, \ldots, \hat{y}_t\}$, where each $\hat{y}_i$ is either 1 or 0. Formally, we have:
\begin{icompact}
\item Step I: EMA. For original prediction $y_t$, the EMA value $s_t$ is computed as $s_t = \frac{\sum_{i=0}^{t} (1 - \alpha)^{t-i} y_i}{\sum_{i=0}^{t} (1 - \alpha)^i}$. We denote $\alpha$ as the parameter that influences the weighting of data points in the EMA calculation.
\item Step II: Majority Vote. The idea is to apply a majority vote to smooth the prediction within a window centered at each EMA value $s_i$ with a padding size $p$. This means that for each $s_i$, we consider its neighboring EMA values within the window and determine the smoothed prediction $\hat{y}_i$ based on the majority of these values being above or below a threshold $\tau$. We define this threshold as the mean of all EMA values: $\tau = \frac{1}{t} \sum_{i=1}^{t} s_i$.
Formally, the smoothed prediction $\hat{y}_i$ is determined as:
\begin{equation}
\hat{y}_i = \begin{cases}
1 & \text{if } \sum_{j=\max(1, i-p)}^{\min(i+p, t)} \mathbbm{1}(s_j > \tau) > \frac{\min(i + p, t) - \max(1, i - p) + 1}{2} \\
0 & \text{otherwise}
\end{cases}
\label{eq:ems}
\end{equation}
where $\mathbbm{1}(\cdot)$ denotes the indicator function and the window size is adaptively defined as $\min(i + p, t) - \max(1, i - p) + 1$ ensuring that the window does not extend beyond the boundaries determined by the range from $\max(1, i - p)$ to $\min(i + p, t)$.
\end{icompact}

\paragraph{Anomaly Score.} Given that $\hat{Y}$ represents the initial detection results of \sys, we can further assess these by calculating an anomaly score through a secondary EMA. Specifically, the \emph{anomaly scores}, denoted as $A = \{a_0, a_1,\ldots, a_t\}$, where $a_t$ is:
\begin{equation}
    a_t = \frac{\sum_{i=0}^{t} (1 - \tau)^{t-i} \hat{y}_i}{\sum_{i=0}^{t} (1 - \tau)^i}
\label{eq:ascore}
\end{equation}
We denote the above procedure \sysbase as a baseline of our method, which provides a \emph{dummy answer}, i.e., ``Anomaly'' if $\hat{y}_i = 1$ otherwise ``Normal'', with an anomaly score that is comparable with the state-of-the-art VAD methods~\cite{aich2023cross_domain_zxVAD,VADRE_liu2021hf2vad,vadfp_morais2019learning,VADRE_shi2023video}. Subsequently, \sys utilizes the dummy answer in the robust reasoning module for further analysis.

\paragraph{Description Modification.}
In this step, \sys modifies the description $D$ comparing $Y$ and $\hat{Y}$ and outputs the modified $\hat{D}$. If $y_i=0$ while $\hat{y}_i=1$, indicating a false negative in the perception module, \sys corrects $d_i$ by adding ``There is a person $\{k\}$.'', where $k\in K$ is the most frequent anomaly keyword within the window size $w$. Conversely, if $y_i=1$ while $\hat{y}_i=0$, indicating a false positive in the perception module, so \sys modifies $d_i$ by removing parts of the description that contain the anomaly keyword $k$. 

\subsection{Robust Reasoning}

In the robust reasoning module, \sys utilizes an LLM to achieve the reasoning task for VAD, with the robust rule $R_{\text{robust}}$ derived from the induction stage as the context. The LLM is fed with each frame's modified description $\hat{d}_i$ with its dummy answer, i.e., either ``Anomaly'' or ``Normal'' generated from \sysbase. We denote the output of robust reasoning as $Y^* = \{\text{LLM}(\hat{d}_i, \hat{y}_i, R_{\text{robust}}, p_r) \mid \hat{d}_i \in \hat{D}, \hat{y}_i \in \hat{Y} \}$. To ensure reliable results, the prompt $p_r$, detailed in Appendix~\ref{appendix:prompts}, guides the LLM to recheck whether the dummy answer $\hat{y}_i$ matches the description $\hat{d}i$ according to $R_{\text{robust}}$. This validation step, instead of directly asking the LLM to analyze $\hat{d}_i$, improves decision-making by using the dummy answer as a hint. This approach helps \sys reduce missed anomalies (false negatives) and ensures that its reasoning is more closely aligned with the rules. Additionally, to compare \sys with the state-of-the-art approaches based on thresholding anomaly scores, we apply Equation~\eqref{eq:ascore} with replacing $\hat{y}_i$ by $y^*_i \in Y^*$ to output anomaly scores.

\section{Experiments}
This section compares \sys with LLM-based baselines and state-of-the-art methods in terms of both detection and reasoning abilities. We also conduct an ablation study on each module within \sys to evaluate their contributions. Examples of complete prompts, derived rules, and outputs are illustrated in Appendix~\ref{appendix:prompts}.

\subsection{Experimental Setup}
\paragraph{Datasets.}
We evaluate our method on four VAD benchmark datasets.
(1) UCSD Ped2 (Ped2)~\cite{li2013anomaly_ped2}: A single-scene dataset captured in pedestrian walkways with over 4,500 frames of videos, including anomalies such as skating and biking.
(2) CUHK Avenue (Ave)~\cite{lu2013abnormal_ave}: A single-scene dataset captured in the CUHK campus avenue with over 30,000 frames of videos, including anomalies such as running and biking.
(3) ShanghaiTech (ShT)~\cite{liu2018future_sht}: A challenging dataset that contains 13 campus scenes with over 317,000 frames of videos, containing anomalies such as biking, fighting, and vehicles in pedestrian areas.
(4) UBnormal (UB)~\cite{acsintoae2022ubnormal}: An open-set virtual dataset generated by the Cinema4D software, which contains 29 scenes with over 236,000 frames of videos.
For each dataset, we use the default training and test sets that adhere to the one-class setting. The normal reference frames used by \sys are randomly sampled from the normal training set. The methods are evaluated on the entire test set if not otherwise specified.

\paragraph{Evaluation Metrics.}
Following the common practice, we use the Area Under the receiver operating characteristic Curve (AUC) as the main detection performance metric. To compare with LLM-based methods that cannot output anomaly scores, we use the accuracy, precision, and recall metrics. Besides, we adopt the Doubly-Right metric~\cite{mao2023doubly} to evaluate reasoning ability. All the metrics are calculated with frame-level ground truth labels.

\paragraph{Implementation Details.}
We implement our method, \sys, using PyTorch~\cite{paszke2019pytorch}. If not otherwise specified, we employ CogVLM-17B~\cite{lvm_wang2023cogvlm} as the VLM for visual perception, GPT-4-1106-Preview~\cite{llm_achiam2023gpt4} as the LLM for induction, and the open-source Mistral-7B-Instruct-v0.2~\cite{llm_jiang2023mistral} as the LLM for deduction (i.e., inference) due to using GPTs on entire test sets is too costly. We discuss other VLMs/LLMs choices in Appendix~\ref{appendix:backbones}. The default hyperparameters of \sys are set as follows: The number of batches in rule aggregation $n = 10$, the number of normal reference frames per batch $m = 1$, the padding size $p =5$ in majority vote, and the weighting parameter $\alpha = 0.33$ in EMA.

\subsection{Comparison with LLM-based Baselines} \label{sec52}
Reasoning for one-class VAD using LLMs is not well-explored. To demonstrate \sys's superiority over the direct LLM use, we build asking LLM/Video-based LLM directly as baselines and also adapt related works~\cite{cao2023towards,elhafsi23_sad} to our target problem as baselines. At test time, let us denote test video frames as $F = \{f_1, f_2, \ldots, f_t\}$. We elaborate on our four baselines as follows. (1) Ask LLM Directly: $\{\text{LLM}(d_i, p) \mid d_i \in D\}$, where the LLM is Mistral-7B, $D$ is $F$'s frame descriptions generated by CogVLM, and $p$ is ``Is this frame description anomaly or normal?'' (2) Ask LLM with Elhafsi et al.~\cite{elhafsi23_sad}: $\{\text{LLM}(d_i, p) \mid d_i \in D\}$, where the LLM is Mistral-7B, $D$ is $F$'s frame descriptions generated by CogVLM, and $p$ is \cite{elhafsi23_sad}'s prompts and predefined concepts of normality/anomaly. (3) Ask Video-based LLMs Directly: $\{\text{Video-based LLM}(c_i, p) \mid c_i \in C\}$, where $p$ is ``Is this clip anomaly or normal?'' We use Video-LLaMA~\cite{lvm_zhang2023videollama} as the Video-based LLM, which performs clip-wise inference. Each video clip $c_i$ consists of consecutive frames in $F$ with the same label. (4) Ask GPT-4V with Cao et al.~\cite{cao2023towards}: $\{\text{GPT-4V}(f_i, p) \mid f_i \in F\}$, where $p$ is \cite{cao2023towards}'s prompts. As a large VLM, GPT-4V directly takes frames as inputs.

\begin{table}[t!]
\centering
\renewcommand{\arraystretch}{1} 
\setlength{\tabcolsep}{17pt}
\scriptsize
\centering
\caption{Detection performance with accuracy, precision, and recall (\%) compared with different VAD with LLM methods on the ShT dataset.}
\label{tab:llm-detect}
\begin{tabular}{l|ccc}
\toprule
\textbf{Method} & \textbf{Accuracy} & \textbf{Precision} & \textbf{Recall} \\
 \midrule
Ask LLM Directly & 52.1 & 97.1 & 6.2 \\
Ask LLM with Elhafsi et al.~\cite{elhafsi23_sad} & 58.4 & \textbf{97.9}& 15.2\\
Ask Video-based LLM Directly &54.7 & 85.4 & 8.5 \\
\hline
\sys & \textbf{81.8} & 90.2 & \textbf{64.3} \\
\bottomrule
\end{tabular}
\end{table}

\paragraph{Detection Performance.}
Table~\ref{tab:llm-detect} compares the accuracy, precision, and recall on the ShT dataset. Overall, \sys achieves significant improvements with an average increase of 26.2\% in accuracy and 54.3\% in recall. Such improvements are attributed to the reasoning based on the rules generated in the induction stage. In contrast, the baselines tend to predict most samples as normal based on the implicit knowledge pre-trained in LLMs, resulting in very low recall and accuracy close to a random guess. Their relatively high precision is due to that they rarely predict anomalies, leading to fewer false positives.

\paragraph{Reasoning Performance.} 
The reasoning performance is evaluated using the Doubly-Right metric~\cite{mao2023doubly}: \{RR, RW, WR, WW\} (\%), where RR denotes Right detection with Right reasoning, RW denotes Right detection with Wrong reasoning, WR denotes Right detection with Wrong reasoning, and WW denotes Wrong detection with Wrong reasoning. We desire a high accuracy of RR (the best is 100\%) and low percentages of RW, WR and WW (the best is 0\%). Since \{RW, WR, WW\} may be caused by visual perception errors rather than reasoning errors, we also consider the case with manually corrected visual perception to exclusively evaluate each method's reasoning ability, i.e., w. Perception Errors vs. w/o. Perception Errors in Table~\ref{tab:llm-reason}.

Due to the lack of benchmarks for evaluating reasoning for VAD, we create a dataset consisting of 100 randomly selected frames from the ShT test set, with an equal split of 50 normal and 50 abnormal frames. For each frame, we offer four choices: one normal and three anomalies, where only one choice with the matched rules is labeled as RR, while the other choices correspond to RW, WR or WW. Details and examples of this dataset are illustrated in Appendix~\ref{appendix:doubly}. Since the 100 randomly selected frames are not consecutive, here \sys's perception smoothing is not used.

Table~\ref{tab:llm-reason} shows the evaluation results. With perception errors, \sys outperforms the baselines by 10\% to 27\% RR, and it achieves a very low WW of 1\% compared to the 17\% WW of the second best Ask GPT-4V with Cao et al.~\cite{cao2023towards}. Without perception errors, \sys's RR jumps to 99\%. These results demonstrate \sys's superiority over the GPT-4(V) baselines and its great ability to make correct detection along with correct reasoning.

\begin{table}[t!]
\centering
\renewcommand{\arraystretch}{1} 
\setlength{\tabcolsep}{6.5pt}
\scriptsize
\centering
\caption{Reasoning performance with the Doubly-Right metric: \{RR, RW, WR, WW\} (\%) on 100 (limited by GPT-4's query capacity) randomly selected frames from the ShT test set. We evaluate cases with visual perception errors (w. Perception Errors) and with manually corrected visual perception (w/o. Perception Errors).}
\label{tab:llm-reason}
\begin{tabular}{l|cccc|cccc}
\toprule
\multirow{2}{*}{\textbf{Method}} & \multicolumn{4}{c|}{\textbf{w. Perception Errors}} & \multicolumn{4}{c}{\textbf{w/o. Perception Errors}} \\
\cline{2-9}
 & RR & RW & WR & WW & RR & RW & WR & WW \\
 \midrule
Ask GPT-4 Directly & 57 & 4 & 15 & 24 & 73 & 3 & 0 & 24 \\
Ask GPT-4 with Elhafsi et al.~\cite{elhafsi23_sad} & 60 & 3 & 15 & 22 & 76 & 2 & 0 & 22 \\
Ask GPT-4V with Cao et al.~\cite{cao2023towards} & 74 & 2 &7 &17 & 81 & 2 &0 &17 \\
\hline
\sys & \textbf{83} & 1 & 15 & 1 & \textbf{99} & 0 & 0 & 1 \\
\bottomrule
\end{tabular}
\end{table}

\subsection{Comparison with State-of-the-Art Methods}
\label{sec:sota}
This section compares \sys with 15 state-of-the-art one-class VAD methods across four datasets, evaluating their detection performance and domain adaptability. The performance values of these methods are sourced from their respective original papers.

\paragraph{Detection Performance.}
Table~\ref{tab:sota} shows the effectiveness of \sys. There are three main observations.
First, \sys, even with its basic version \sysbase, outperforms all the Image-Only competitors, which also do not use any additional features (e.g., bounding boxes from object detectors or 3D features from action recognition networks), on the challenging ShT and UB datasets. This suggests that our rule-based reasoning benefits the challenging one-class VAD task.
Second, for Ped2 and Ave, \sys performs on par with the Image-Only methods. This is achieved without any tuning, meaning that our few-normal-shot prompting approach is as effective as the costly full-shot training on these benchmarks. Third, \sys outperforms \sysbase, indicating that the robust reasoning module improves performance further.

\begin{table}[t!]
\centering
\renewcommand{\arraystretch}{1} 
\setlength{\tabcolsep}{5.5pt}
\scriptsize
\centering
\caption{AUC (\%) compared with different one-class VAD methods. ``Image Only'' methods only rely on image features. In contrast, others employ additional features such as bounding boxes from object detectors or 3D features from action recognition networks. ``Training'' indicates the methods that need a full-shot training process.}
\vspace{-0.1in}
\label{tab:sota}
\begin{tabular}{l|l|c|c|c|c|c|c}
\toprule
\textbf{Method} & \textbf{Venue} & \textbf{Image Only} & \textbf{Training} & \textbf{Ped2} & \textbf{Ave} & \textbf{ShT} & \textbf{UB} \\
\midrule

MNAD~\cite{park2020learning_memg}          &  CVPR-20       & \cmark            &   \cmark         & 97.0            &   88.5     &  70.5      & -           \\ 

rGAN~\cite{lu2020few_domain_rgan}      & ECCV-20        & \cmark             &  \cmark           & 96.2       &  85.8    &  77.9      & -           \\
CDAE~\cite{chang2020clustering_cdae}            &  ECCV-20        &  \cmark           &  \cmark       & 96.5            & 86.0       & 73.3       & -           \\

MPN~\cite{lv2021learning_domain_mpn}         & CVPR-21        & \cmark             &  \cmark         &     96.9       &   89.5    &  73.8     & -           \\ 
NGOF~\cite{wang2021video_ngof}          &  CVPR-21       & \xmark            &  \cmark          & 94.2            &   88.4     &  75.3     & -           \\ 
HF2~\cite{VADRE_liu2021hf2vad}          & ICCV-21        & \xmark             &  \cmark         & \textbf{99.2}            &  91.1      & 76.2       & -           \\ 
BAF~\cite{VADDI_georgescu2021background}          & TPAMI-21        &  \xmark           & \cmark           & 98.7       & 92.3       & 82.7       & 59.3           \\

GCL~\cite{zaheer2022generative_gcl}          & CVPR-22        & \xmark             &  \cmark           & -            & -       &  79.6      & -           \\ 
S3R~\cite{wu2022self}          & ECCV-22        & \xmark             &  \cmark           & -            &    -    &  80.5      & -           \\
SSL~\cite{VADRE_wang2022video}          & ECCV-22        & \xmark             &  \cmark           & 99.0            & \textbf{92.2}       &  84.3      & -           \\
 
zxVAD~\cite{aich2023cross_domain_zxVAD}          & WACV-23        & \xmark             &  \cmark           & 96.9            &   -   &    71.6   & -  \\
HSC~\cite{VADRE_sun2023_scene}       & CVPR-23    & \xmark                    &\cmark                  &  98.1                & 93.7            & 83.4       & -           \\
FPDM~\cite{VADRE_Yan_2023_ICCV}          &ICCV-23         &   \cmark          &  \cmark          & -            & 90.1       & 78.6       & 62.7           \\ 
SLM~\cite{VADRE_shi2023video}            & ICCV-23         &  \cmark          &  \cmark                 & 97.6            & 90.9       & 78.8       & -           \\
{STG-NF}~\cite{hirschorn2023normalizing} & ICCV-23 & \xmark & \cmark & - & - &\textbf{85.9} &71.8 \\
\hline
\sysbase           &  -       &  \cmark          & \xmark         & 96.5            & 82.2       & 84.6       & 69.8          \\
\sys           &  -       &  \cmark          &  \xmark            &  97.9           & 89.7           &   85.2    &    \textbf{71.9}           \\

\bottomrule
\end{tabular}
\end{table}

\begin{table}[t!]
\centering
\renewcommand{\arraystretch}{1} 
\setlength{\tabcolsep}{5.5pt}
\scriptsize
\centering
\caption{AUC (\%) compared with different cross-domain VAD methods. We follow the compared works to use ShT as the source domain dataset for other target datasets.}
\label{tab:domain}
\begin{threeparttable}
\begin{tabular}{l|l|c|c|c|c|c|c}
\toprule
\textbf{Method} & \textbf{Venue} & \textbf{Image Only} & \textbf{Training} & \textbf{Ped2} & \textbf{Ave} & \textbf{ShT\tnote{1}} & \textbf{UB} \\
\midrule

rGAN~\cite{lu2020few_domain_rgan}          & ECCV-20        & \cmark             &  \cmark           & 81.9       &  71.4    &  77.9      & -           \\
MPN~\cite{lv2021learning_domain_mpn}          & CVPR-21        & \cmark             &  \cmark         &     84.7       &   74.1    &  73.8     & -           \\ 
zxVAD~\cite{aich2023cross_domain_zxVAD}          & WACV-23        & \xmark             &  \cmark           & 95.7            &   \textbf{82.2}   &    71.6   & -  \\
\hline
\sysbase            &  -       &  \cmark          & \xmark         & \textbf{97.4}            & 81.6       & \textbf{83.5}       & \textbf{65.4}           \\
\bottomrule
\end{tabular}
\begin{tablenotes}
\item[1] \sys employs UB as the source domain when ShT serves as the target domain. The competitors have no cross-domain evaluation on ShT, so we report their same-domain results.
\end{tablenotes}
\end{threeparttable}
\end{table}

\paragraph{Domain Adaptability.}
Domain adaptation considers the scenario that the source domain (i.e., training/induction) dataset differs from the target domain (i.e., testing/deduction) dataset~\cite{ganin2015unsupervised,lo2023spatio,tsai2018learning}. We compare \sys with three state-of-the-art VAD methods that claim their domain adaptation ability ~\cite{aich2023cross_domain_zxVAD,lu2020few_domain_rgan,lv2021learning_domain_mpn}. We follow the compared works to use ShT as the source domain dataset for other target datasets. As shown in Table~\ref{tab:domain}, \sys achieves the highest AUC on Ped2, ShT and UB, outperforming with an average of 9.88\%. While \sys trails zxVAD~\cite{aich2023cross_domain_zxVAD} by 0.6\%, it is still higher than the others with an average of 8.85\%. The results indicate that \sys has better domain adaptability across different datasets. This advantage is due to that the language provides consistent descriptions across different visual domains, which allows the application of induced rules to datasets with similar anomaly scenarios but distinct visual appearances. In contrast, traditional methods extract high-dimensional visual features that are sensitive to visual appearances, thereby struggling to transfer their knowledge across datasets.

\subsection{Ablation Study}\label{sec:ab}

In this section, we look into how the proposed strategies affect \sys. We investigate two aspects: rule quantity (i.e., the number of induced rules) and rule quality (i.e., their resulting performance). Regarding this, we evaluate variants of \sysbase on the ShT dataset. 

\paragraph{Ablation on Strategies.}
Table~\ref{tab:ab-1} shows the effects of removing individual strategies compared to using all strategies. In terms of rule quantity, removing Human and Environment or Normal and Anomaly significantly reduces rules by 47.6\% and 82.4\%, respectively. This reduction is due to not separating the rules for humans and the environment halves the number of rules. Moreover, without deriving anomaly rules from normal rules, we only have a limited set of normal rules. Removing Abstract and Concrete or Rule Aggregation slightly increases the number of rules, as the former merges rules within the same categories and the latter removes incorrect rules. Perception Smoothing does not affect rule quantity since it is used in the deduction stage.
In terms of rule quality, removing Normal and Anomaly or Rule Aggregation has the most negative impact. The former happens because when only normal rules are present, the LLM overreacts to slightly different actions such as ``walking with an umbrella'' compared to the rule for ``walking'', leading to false positives. Furthermore, without rules for anomalies as a reference, the LLM easily misses anomalies. The latter is due to that perception errors in the induction stage would lead to incorrect rules for normal.
Besides, removing other strategies also decreases AUC, underscoring their significance. In summary, the proposed strategies effectively improve \sys's performance. There is no direct positive/negative correlation between rule quantity and quality, i.e., having too few rules leads to inadequate coverage of normality and anomaly concepts while having too many rules would cause redundancy and errors.

\begin{figure*}[!t]
\centering
\subcaptionbox{$n$ \& $m$ vs. \# Rules\label{fig:ab-2-1}}{\includegraphics[width=0.31\linewidth]{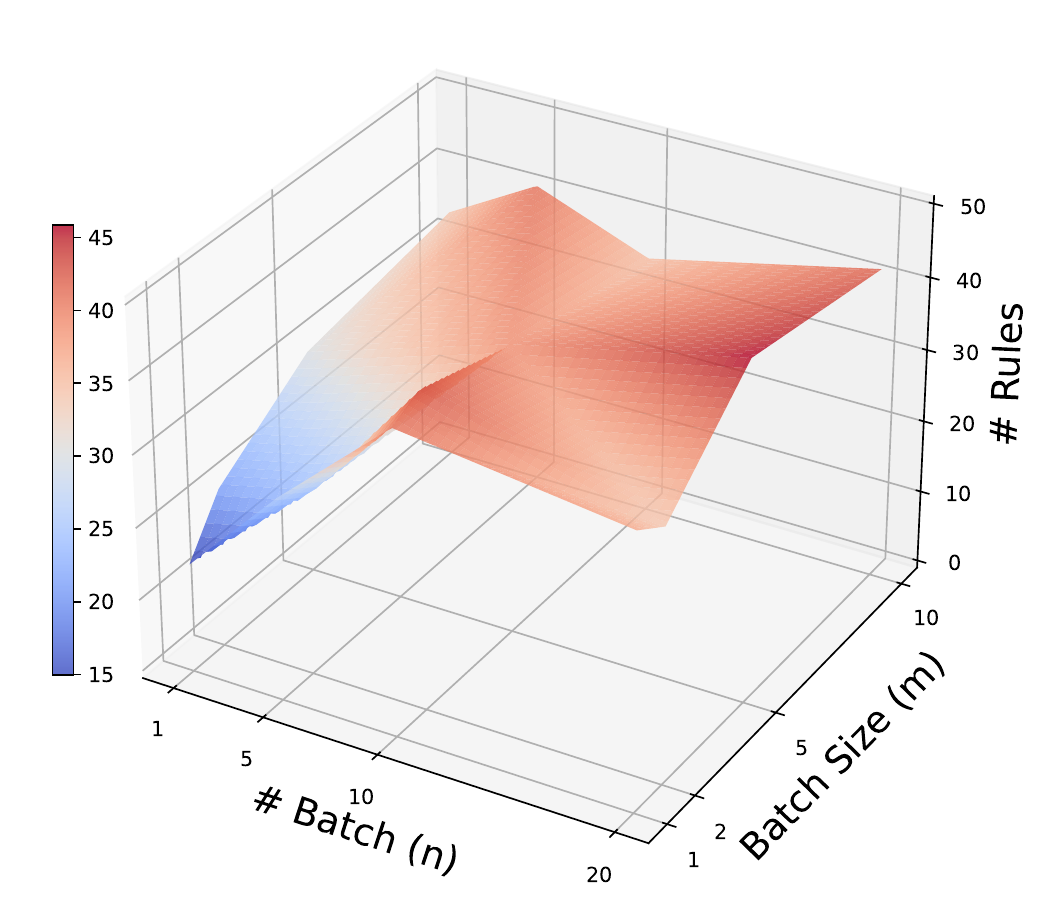}}
\subcaptionbox{$n$ \& $m$ vs. AUC\label{fig:ab-2-2}}{\includegraphics[width=0.31\linewidth]{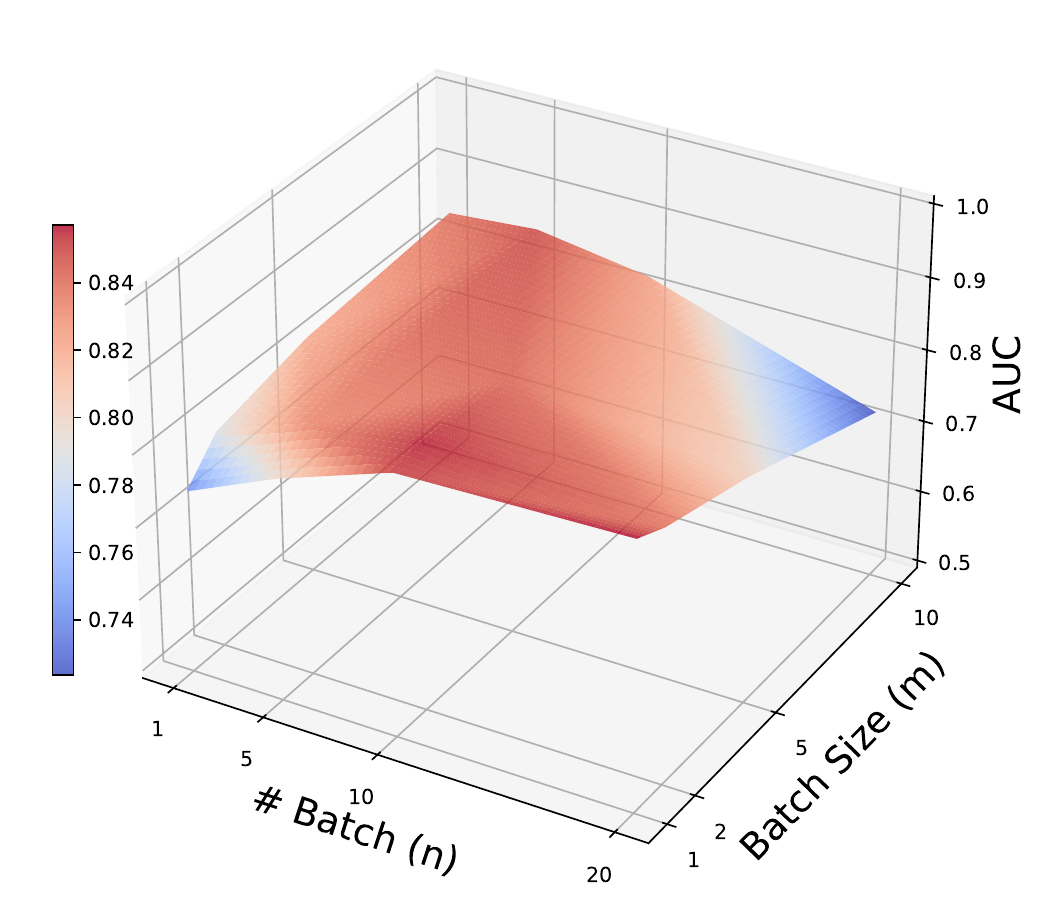}}
\subcaptionbox{$p$ \& $\alpha$ vs. AUC \label{fig:ab-2-3}}{\includegraphics[width=0.31\linewidth]{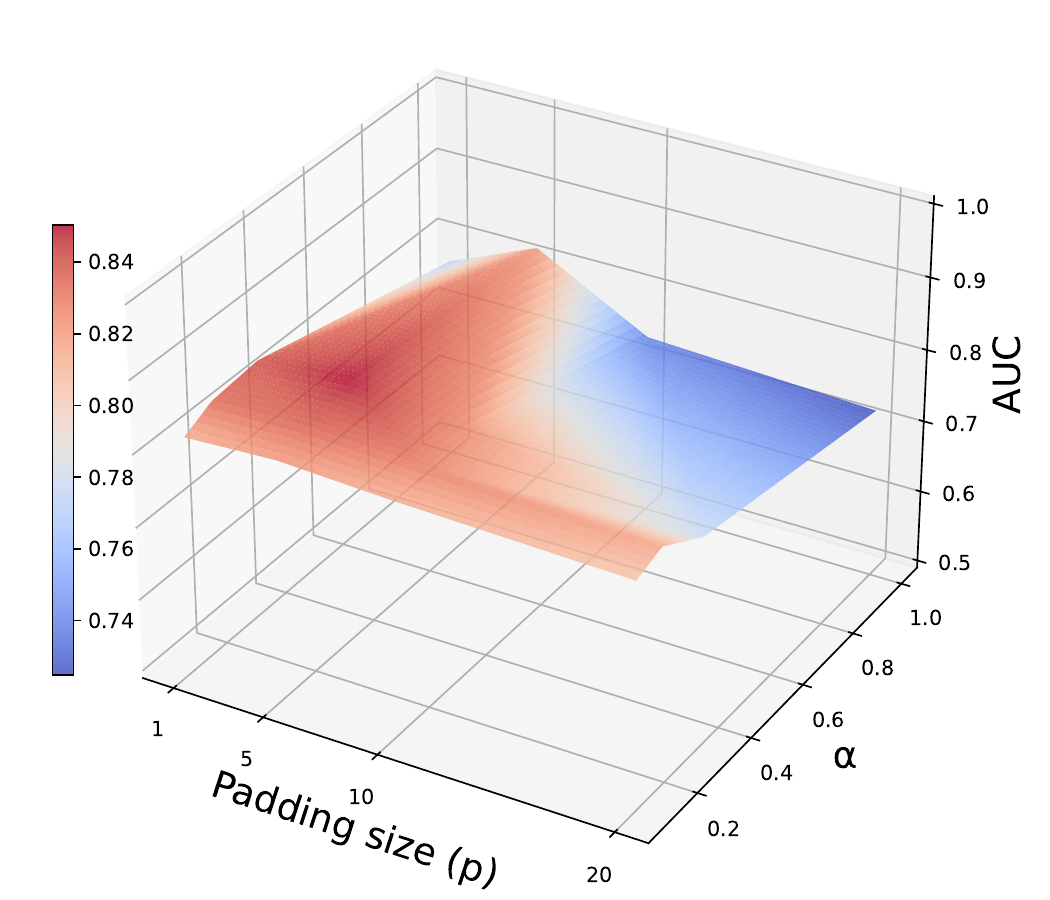}}
\caption{Ablation on hyperparameters of the  (a) (b) rule aggregation and (c) perception smoothing modules on the ShT dataset.}  
\label{fig:ab-2}
\end{figure*}

\paragraph{Ablation on Hyperparameters.}
Fig.~\ref{fig:ab-2} illustrates the effects of the hyperparameters in the rule aggregation and perception smoothing modules. For rule aggregation, we conduct cross-validation on the number of batches $n$ = [1, 5, 10, 20] and the number of normal reference frames per batch $m$ = [1, 2, 5, 10]. We observe that both the number of rules and AUC increase with the increases of $n$ and $m$, but they start to fluctuate when $n \times m$ becomes large. For example, when $n=20$, AUC drops from 85.9\% to 72.2\% as $m$ increases because having too many reference frames (e.g., over 100) results in redundant information in a long context. For perception smoothing, we test the padding size in majority vote $p$ = [1, 5, 10, 20] and the weighting parameter in EMA $\alpha$ = [0.09, 0.18, 0.33, 1]. We found $p=5$ to be optimal for capturing the motion continuity in a video while avoiding the excessive noise that can occur with more neighborhoods. $\alpha$ adjusts the weight of the most recent frames compared to previous frames. A smaller $\alpha$ emphasizes previous frames, resulting in more smoothing but less responsiveness to recent changes. In general, increasing $\alpha$ from 0.09 to 0.33 improves AUC, suggesting that moderate EMA smoothing is beneficial.

\begin{table}[t!]
\centering
\renewcommand{\arraystretch}{1} 
\setlength{\tabcolsep}{0.7pt}
\scriptsize
\centering
\begin{threeparttable}
\caption{Ablation on strategies. We assess the effects of removing individual strategies in \sys. We conduct the experiments five times with different randomly selected normal reference frames for induction and report their mean and standard deviation on the ShT dataset.}
\label{tab:ab-1}
\begin{tabular}{l|c|cc|cc|cc|cc|cc}
\toprule
\multirow{2}{*}{\textbf{Strategy}} & \multirow{2}{*}{\textbf{Stage}} & \multicolumn{2}{c|}{\textbf{\# Rules}} & \multicolumn{2}{c|}{\textbf{Accuracy}} & \multicolumn{2}{c|}{\textbf{Precision}} & \multicolumn{2}{c|}{\textbf{Recall}} & \multicolumn{2}{c}{\textbf{AUC}}  \\
\cline{3-12}
& & mean & std  & mean & std  & mean & std  & mean & std  & mean & std \\
\midrule
w. All Below (default) & Both  & 42.2  & 4.2 & 81.6 & 1.3  &90.9 & 0.8  & 63.9 & 2.7  & 84.5 & 1.1 \\
\hline
w/o. Human and Environment   & Both & -20.1 & +1.1 & -3.3 &+0.8 & -3.9 & +0.8 & -1.9 & +1.6 &-2.4 & +2.0  \\ 
w/o. Normal and Anomaly & Induction & -34.8 & -1.3 & -20.5 & +4.3 & -41.2 & +7.0 & -14.4 &+11.6 &-18.8 & +1.2  \\
w/o. Abstract and Concrete & Induction & +2.3 & +2.7 & -0.6 & -0.2 & -0.9 & -0.2 & -0.3 & -0.4 &-0.9 &+0.1 \\
w/o. Rule Aggregation & Induction & +8.5 & +6.1 & -9.6 & + 14.7 &+1.1 & +2.9 & -10.7 &+14.1 &-15.8 &+0.8    \\
w/o. Perception Smoothing & Deduction & NA & NA & -1.7 & -0.9 &-1.9 &+0.1 &-3.8 &-0.3 &-3.3 &+0.8  \\
\bottomrule
\end{tabular}
\end{threeparttable}
\end{table}

\section{Conclusion}
In this paper, we propose \sys, a novel rule-based reasoning framework for VAD with LLMs. With the induction and deduction stages, \sys requires only few-normal-shot prompting without the need for expensive full-shot tuning, thereby fast steering LLMs' reasoning strengths to various specific VAD applications. To the best of our knowledge, \sys is the first reasoning approach for one-class VAD. Extensive experiments demonstrate \sys's state-of-the-art performance, reasoning ability, and domain adaptability. Limitations and potential negative social impact of this work are discussed in the Appendix~\ref{appendix:limitations}. In future research, we expect this work to advance broader one-class problems and related tasks, such as industrial anomaly detection~\cite{bergmann2019mvtec,you2022unified}, open-set recognition~\cite{bendale2016towards,safaei2023open}, and out-of-distribution detection~\cite{hendrycks2017baseline,sharifi2024gradient}.

\section*{Acknowledgments}

This work was supported in part by National Science Foundation (NSF) under grants OAC-23-19742 and Johns Hopkins University Institute for Assured Autonomy (IAA) with grants 80052272 and 80052273. The views and conclusions contained herein are those of the authors and should not be interpreted as necessarily representing the official policies or endorsements, either expressed or implied, of NSF or JHU-IAA.

\bibliographystyle{splncs04}
\bibliography{References}

\newpage
\appendix
\onecolumn
\section{Appendix}

\subsection{Limitations and Potential Negative Social Impact}
\label{appendix:limitations}

\paragraph{Limitations.}
Similar to most existing LLM-based studies, \sys assumes that the employed LLM backbones have decent capabilities. Sub-optimal LLMs may hinder the effectiveness of the methods. Exploring this limitation further could be an interesting future investigation.

\paragraph{Potential Negative Social Impact.}
The proposed method may enable malicious actors to more easily adapt VLMs/LLMs for illegal surveillance. To mitigate this risk, computer security mechanisms could be integrated.

\subsection{Examples of Input Prompts and Outputs Results}
\label{appendix:prompts}

\paragraph{Induction.} This stage starts from $n$ randomly chosen normal reference frames $F_{\text{normal}} = \{f_{\text{normal}_1}, \text{ } \ldots, \text{ } f_{\text{normal}_n}\}$ and outputs a set of robust rules $R_\text{robust}$. To simplify the illustration, we show one frame $f_{\text{normal}_i} \in F_{\text{normal}}$ as an example in the visual perception and rule generation steps.

\begin{icompact}
\item Visual Perception
\begin{itemize}
\item Input $f_{\text{normal}_i}$ and prompt $p_v$:

\scriptsize
\begin{minipage}{0.17\textwidth}
    \includegraphics[width=\linewidth]{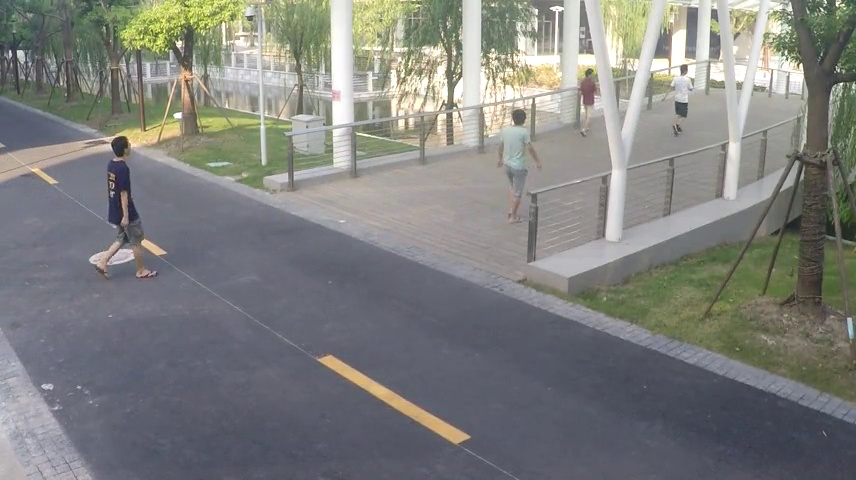}
\end{minipage}%
\hfill
\begin{minipage}{0.72\textwidth}
\begin{lstlisting}
$p_v$ = How many people are in the image and what is each of them doing? What are in the image other than people? Think step by step.
\end{lstlisting} 
\end{minipage}

\normalsize
\item Outputs: Frame description $d_{\text{normal}_i} = VLM(f_{\text{normal}_i}, p_v)$

\scriptsize
\begin{lstlisting}
$d_{\text{normal}_i}$ = There are four people in the image. Starting from the left, the first person is walking on the path. The second person is walking on the bridge. The third person is also walking on the bridge. The fourth person is also walking on the bridge. Other than people, there are trees, a railing, a path, and a bridge visible in the image.
\end{lstlisting} 
\end{itemize}

\item Rule Generation
\begin{itemize}
\item Input prompt $p_g$:
\begin{lstlisting}
$p_g$ = [
    {``role'': ``system'',
     ``content'': As a surveillance monitor for urban safety using the ShanghaiTech dataset, my job is to derive rules for detecting abnormal human activities or environmental objects.},
    {``role'': ``user'', 
     ``content'': Based on the assumption that the given frame descriptions are normal, Please derive rules for normal, start from an abstract concept, and then generalize to concrete activities or objects.},
    {``role'': ``assistant'', 
     ``content'':
                **Rules for Normal Human Activities:
                1. 
                **Rules for Normal Environmental Objects:
                1.
                },
    {``role'': ``user'',
     ``content'': Compared with the above rules for normal, can you provide potential rules for anomaly? Please start from an abstract concept then generalize to concrete activities or objects, compared with normal ones.},
    {``role'': ``assistant'', 
     ``content'':
                **Rules for Anomaly Human Activities:
                1. 
                **Rules for Anomaly Environmental Objects:
                1.
                },
    
    {``role'': ``user'',
     ``content'': Now you are given frame description {$d_{\text{normal}_i}$}. What are the Normal and Anomaly rules you have? Think step by step. Reply following the above format, start from an abstract concept and then generalize to concrete activities or objects. List them using short terms, not an entire sentence.},
    ]
\end{lstlisting}
\item Outputs: For each normal reference frame $d_{\text{normal}_i}$, we will get one set of rules $r_i = LLM(d_{\text{normal}_i}, p_g)$.
Since the structure of the rules is identical to the robust rules, we only present the robust rules in the following step as an illustration of our final induction output.
\end{itemize}

\item Rule Aggregation
\begin{itemize}
\item Input prompt $p_a$:
\begin{lstlisting}
$p_a$ = [
    {``role'': ``system'',
     ``content'': As a surveillance monitor for urban safety using the ShanghaiTech dataset, my job is to organize rules for detecting abnormal activities and objects.},
    {``role'': ``user'', 
     ``content'': You are given {$n$} independent sets of rules for Normal and Anomaly. For the organized normal Rules, list the given normal rules with high-frequency elements For the organized anomaly Rules, list all the given anomaly rules},
    {``role'': ``assistant'', 
     ``content'':
                **Rules for Anomaly Human Activities:
                1. 
                **Rules for Anomaly Environmental Objects:
                1. 
                **Rules for Normal Human Activities:
                1. 
                **Rules for Normal Environmental Objects:
                1. 
                },
    {``role'': ``user'',
     ``content'': Now you are given {$n$} independent sets of rules as the sublists of {$R$}. What rules for Anomaly and Normal do you get? Think step by step, and reply following the above format.},
    ]
\end{lstlisting}
\item Outputs: Robust rules $R_\text{{robust}} = LLM(R = \{r_1, \text{ } \ldots, \text{ } r_n\}, p_a)$
\begin{lstlisting}
**Rules for Anomaly Human Activities:**
1. Using any non-walking movement such as riding a bicycle or scooter, skateboarding
2. Running or jumping
3. Moving with the usual speed such as very fast or in a hurry
4. Person lying or bending down on the ground
5. Aggressive or unusual movements such as fighting or pushing
6. Loitering around the trash bin
7. Climbing on trees or into manhole/utility access points
8. Tampering with manhole covers
9. Engaging in physical altercation
10. Lingering without a clear purpose near a sensitive area
11. Falling unexpectedly

**Rules for Anomaly Environmental Objects:**
1. Missing, removed, or ajar manhole covers
2. Ground transportation such as vehicles, vans, bicycles.
3. Unattended bags or objects (security threat)
4. Unusual objects (vehicles, large unattended luggage, etc.)
5. Unauthorized posters or signs
6. Damaged manhole cover or trees
7. Abandoned items in a public area
8. Yellow objects (when not commonly seen)
9. Graffiti on building windows or vandalism to a fence
10. Obstructions on pedestrian crossing
11. Fire hydrant spraying water without presence of emergency services
12. Unidentified objects near the tall structure
13. Smoke or fire coming from a utility access point
14. Objects thrown or falling from a height

**Rules for Normal Human Activities:**
1. Walking alone or with another person
2. Standing and using a mobile device
3. Walking on the sidewalk
4. Walking towards or away from a building
5. Entering a building
6. Standing near a trash bin
7. Waiting at traffic lights
8. Walking on pedestrian crossing
9. Sitting on bench or ground

**Rules for Normal Environmental Objects:**
1. Manhole covers on the ground or street
2. Trees along the street or Plants present
3. Trash bin in vicinity or placed beside the street
4. Posters on glass door
5. Static building with glass windows
6. Fence along the water
7. Pedestrian crossing markings visible
8. Standing fire hydrant
9. Static tall structure in the background
10. Utility access points on the ground
\end{lstlisting}
\end{itemize}
\end{icompact}

\paragraph{Deduction.} This stage starts from a test video that contains $t$ continuous frames $F = \{f_1, \text{ } \ldots, \text{ } f_t\}$ and outputs the reasoning results $\hat{Y}^* = \{\hat{y}^*_1, \text{ } \ldots, \text{ } \hat{y}^*_t\}$. To simplify the illustration, we show two frames of this test video, $f_{i}, f_j  \in F$ as examples that represent one anomaly frame and one normal frame, respectively.
\begin{icompact}
\item Visual Perception:  
\begin{itemize}
\item Input test frames $f_i$, $f_j$ and prompt $p_v$:

\begin{minipage}{0.17\textwidth}
    \includegraphics[width=\linewidth]{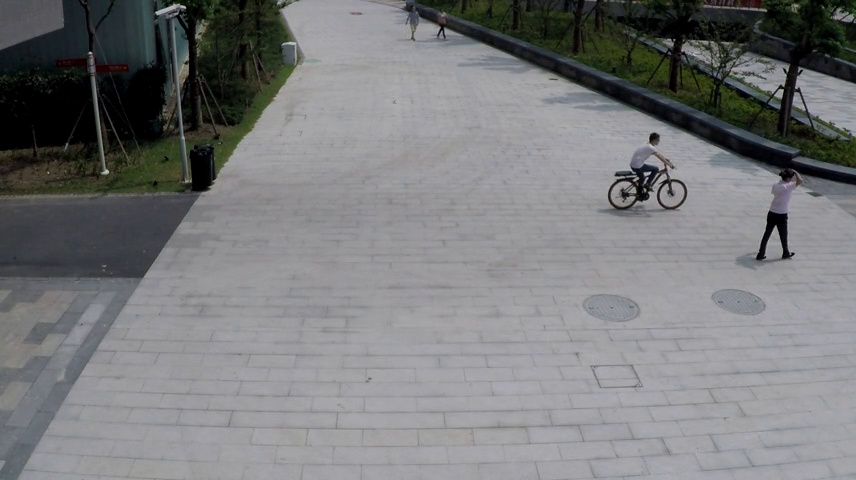}
\end{minipage}%
\hfill
\begin{minipage}{0.72\textwidth}
\begin{lstlisting}
$p_v$ = How many people are in the image and what is each of them doing? What are in the image other than people? Think step by step.
\end{lstlisting} 
\end{minipage}

\begin{minipage}{0.17\textwidth}
    \includegraphics[width=\linewidth]{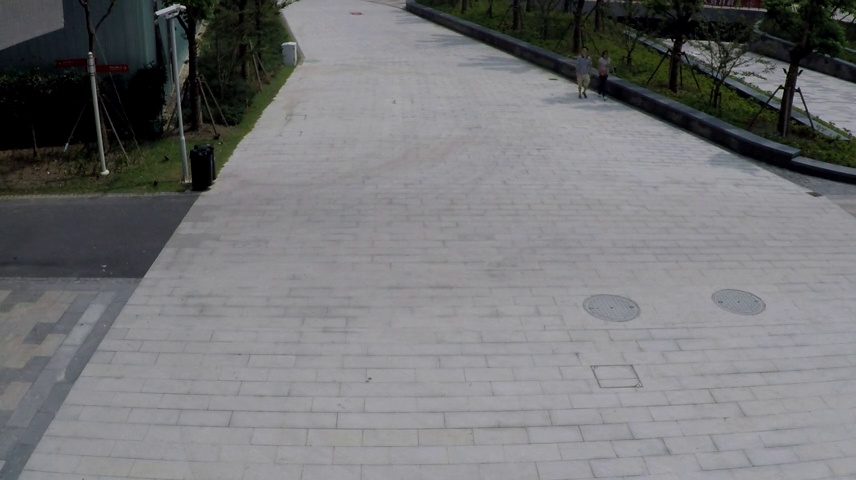}
\end{minipage}%
\hfill
\begin{minipage}{0.72\textwidth}
\begin{lstlisting}
$p_v$ = How many people are in the image and what is each of them doing? What are in the image other than people? Think step by step.
\end{lstlisting} 
\end{minipage}

\item Outputs: Frame descriptions $d_i = VLM(f_i, p_v)$, $d_j = VLM(f_j, p_v)$
\begin{lstlisting}
$d_i$ = There are four people in the image. One person is walking, another is also walking, the third person is riding a bicycle, and the fourth person is walking near the bicycle. Other than people, there are trees, a pathway, a trash bin, a bicycle, and two manhole covers visible in the image.
\end{lstlisting}
\begin{lstlisting}
$d_j$ = There are two people in the image. One person appears to be walking, the other seems to be walking together. Other than people, there are two manhole covers on the ground, a trash bin, and some trees and plants.
\end{lstlisting}
\end{itemize}

\item Perception Smoothing:
\begin{itemize}
\item $R_\text{robust} \rightarrow K$ (generate anomaly keywords from anomaly rules, see Section \ref{sec:de}).
    \begin{itemize}
        \item Input prompt $p_k$:
        \begin{lstlisting}
$p_k$ = You will be given a set of rules for detecting abnormal activities and objects; please extract the anomaly keywords, activities using ``ing'' verbs, and anomaly objects using nouns, and provide a combined Python list with each represented by a single word. The output should be in the format: ["object1", ..., "activity1", "activity2", ...]. Now you are given $\{R_\text{robust}\}:$
        \end{lstlisting}
        \item Output $K$:
        \begin{lstlisting}
anomaly_from_rule = ["trolley", "cart", "luggage", "bicycle", "skateboard", "scooter", "vehicles", "vans", "accident", "running", "jumping", "riding", "skateboarding", "scooting", "lying", "falling", "bending", "fighting", "pushing", "loitering", "climbing", "tampering", "lingering"]
        \end{lstlisting}
    \end{itemize}
\item $d_i \rightarrow \hat{d}_i \,\&\, \hat{y}_i$, $d_j \rightarrow \hat{d}_j \,\&\, \hat{y}_j$ (generate modified descriptions and smoothed predictions, see Section \ref{sec:de}).
\end{itemize}

\item Robust Reasoning:
\begin{itemize}
\item Input prompt $p_r$:
\begin{lstlisting}
$p_r$ = You will be given an description of frame, your task is to double check my initial anomaly detection result based on the rules. 
    The rules are: {$R_\text{robust}$}
    My initial result is {$\hat{y}_i$}
    First, if human activity is present, which rule is matching? List the rule category, e.g., normal or anomaly, with the rule number.
    Second, if an environmental object is present, which rule is matching? List the rule category, e.g., normal or anomaly,  with the rule number.
    Third, are the human activities or environmental objects anomaly? Answer: anomaly, if ANY anomaly rule (even if only one, no matter human activities or environmental objects) matches, otherwise answer: normal.
    Now you are given the frame {$\hat{d}_i$}, think step by step.
\end{lstlisting}
\item Final reasoning output: $y^*_i = LLM(\hat{d}_i, \hat{y}_i, R_\text{{robust}}, p_r)$
\begin{lstlisting}
$\hat{y}^*_i$ = The third person is riding a bicycle, which matches rule number 1 under Anomaly Human Activities.

     The bicycle matches rule number 2 under Anomaly Environmental Objects.

     Answer: anomaly.
\end{lstlisting}
$y^*_j = LLM(\hat{d}_j, \hat{y}_j, R_\text{{robust}}, p_r)$
\begin{lstlisting}
$\hat{y}^*_j$ = The human activities present are two people walking. The rules for normal human activities include walking alone or with another person. So, the rule matching is:
    1. Walking alone or with another person (normal)

    The environmental objects present are two manhole covers on the ground, a trash bin, and some trees and plants. So, the rules matching are:
    1. Manhole covers on the ground or street (normal)
    2. Trees along the street or Plants present (normal)
    3. Trash bin in vicinity or placed beside the street (normal)

    Answer: normal.
\end{lstlisting}
\end{itemize}
\end{icompact}

\subsection{Examples of Doubly-Right Reasoning Evaluation}
\label{appendix:doubly}

In Section \ref{sec52}, we evaluate reasoning ability using the Doubly-Right metric~\cite{mao2023doubly}. We create a benchmark dataset with multiple choices for Doubly-Right reasoning evaluation. The evaluation is conducted in the deduction stage, where we input the visual perception description and the induced rules to an LLM. The goal is to demonstrate that the induced rules enable LLMs to perform correct reasoning.

We list the prompt for reasoning evaluation below and one example of the description and its four choices as Table~\ref{tab:reasoning}. The content in normal choice is fixed, while the anomaly choices include one correct reasoning with a matched rule and two randomly chosen non-matched rules from our generated anomaly rules. In this example, Choices A, B, C and D correspond to RW, WW, RR and RW, respectively.

\begin{lstlisting}
[
{``role'': ``system'',
 ``content'': You will be given a description of the frame and four choices. Your task is to make the correct choice based on the rules. The rules are: {$R_\text{robust}$}},
{``role'': ``user'',
 ``content'': Description: {$\hat{d}_i$}
Choices: {Four Choices}
Choose just one correct answer from the options (A, B, C, or D) and output without any explanation. Please Answer:},
]
\end{lstlisting}
\begin{table}[ht]
\centering
\scriptsize
\caption{An example of reasoning performance evaluation with multiple reasoning choices. In this example, Choices A, B, C and D correspond to RW, WW, RR and RW of the Doubly-Right metric, respectively. The RR choice is highlighted in \highlight{yellow}.}
\label{tab:reasoning}
\begin{tabular}{|p{0.32\textwidth}|p{0.66\textwidth}|}
\hline
\textbf{Frame Description} & \textbf{Multiple Choices for Reasoning Evaluation} \\ \hline
There are four people in the image. One person is walking with a backpack, another person is riding a bicycle, a third person is standing and looking at the bicyclist, and the fourth person is sitting on a bench. Other than people, there are trees, a trash bin, and two manhole covers visible in the image. &
A. Anomaly, since ``climbing on a tree'' matches anomaly human activities ``Climbing on trees or into manhole/utility access points''. \newline
B. Normal, since no rules for anomaly human activities or non-human objects match. \newline
\highlight{C. Anomaly, since ``riding a bicycle'' matches} anomaly human activities ``Using any non-walking movement such as riding a bicycle or scooter, skateboarding''. \newline
D. Anomaly, since ``a vehicle parked blocking a pedestrian crossing'' matches anomaly non-human objects ``Obstructions on pedestrian crossing''. \\ \hline
\end{tabular}
\end{table}

\subsection{Different VLMs/LLMs as Backbones}
\label{appendix:backbones}
Table~\ref{tab:differentllms} shows the results of using various VLMs/LLMs as backbones in the deduction stage, compared to the default setting (the first row). All the results are based on the same rules derived in the induction stage with the default setting.

\begin{table}[!ht]
\centering
\renewcommand{\arraystretch}{1.2} 
\setlength{\tabcolsep}{3.5pt}
\scriptsize
\centering
\caption{Detection performance with accuracy, precision, and recall (\%) using different VLMs/LLMs as backbones in the deduction stage on 100 (limited by GPT-4’s query capacity) randomly selected frames from the ShT test set.} 
\label{tab:differentllms}
\begin{tabular}{l|l|ccc|c}
\toprule
\textbf{Visual Perception} & \textbf{Robust Reasoning} & \textbf{Accuracy} & \textbf{Precision} & \textbf{Recall}&  \textbf{Open Source} \\
\midrule
CogVLM~\cite{lvm_wang2023cogvlm} (default) & Mistral~\cite{llm_jiang2023mistral} (default) & 82.0 & 88.1 & 74.0 & \cmark  \\
\hline
GPT-4V~\cite{llm_achiam2023gpt4} & GPT-4V & 83.0 & 88.4 & 76.0 & \xmark \\
LLaVA~\cite{lvm_liu2023llava} & LLaVA  & 40.0 & 40.4 & 42.0 & \cmark \\
PandaGPT~\cite{su2023pandagpt} & PandaGPT & 37.0 & 31.4 & 22.0  & \cmark  \\
\hline
OWLViT~\cite{minderer2022simple} & \multirow{4}{*}{Mistral} & 71.0 & 82.0 & 54.0 & \cmark \\
LLaVA & & 76.0 & 79.5 & 70.0 & \cmark \\
BLIP-2~\cite{lvm_li2023blip2} & & 50.0 & 50.0 & 94.0 & \cmark  \\
RAM~\cite{zhang2023recognize} & & 45.0 & 47.2 & 84.0 & \cmark \\
\hline
\multirow{2}{*}{CogVLM} & GPT-3.5~\cite{brown2020language} & 81.0 & 86.0 & 74.0 & \xmark \\
& LLaMA-2~\cite{llm_touvron2023llama2} & 60.0 & 70.8 & 34.0 & \cmark \\
\bottomrule
\end{tabular}
\end{table}

We categorize the comparisons into three types: (1) VLMs only: \sys uses the same VLMs as an end-to-end solution, combining visual perception and robust reasoning. It inputs the test frame and outputs the reasoning result. This category includes GPT-4V~\cite{llm_achiam2023gpt4}, LLaVA~\cite{lvm_liu2023llava}, and PandaGPT~\cite{su2023pandagpt}. (2) VLMs + Mistrial~\cite{llm_jiang2023mistral}: We keep Mistrial as the default LLM for robust reasoning and test different VLMs (e.g., OWLViT~\cite{minderer2022simple}, LLaVA, BLIP-2~\cite{lvm_li2023blip2}, RAM~\cite{zhang2023recognize}) for visual perception. (3) CogVLM~\cite{lvm_wang2023cogvlm} + LLMs: We use CogVLM as the fixed VLM for visual perception and test different LLMs for robust reasoning (e.g., GPT-3.5~\cite{brown2020language}, LLaMA-2~\cite{llm_touvron2023llama2}). We have the following observations.

For the VLMs-only category, GPT-4V performs well, but it has limitations on the number of queries and a high cost per query, making it expensive for large-scale testing. LLaVA and PandaGPT, on the other hand, show poor reasoning ability. They cannot follow the provided robust rules, and generate irrelevant content or hallucinations. An example frame with their outputs is shown below:

\noindent
\fbox{
\scriptsize
\begin{minipage}{0.25\textwidth}
    \includegraphics[width=\linewidth]{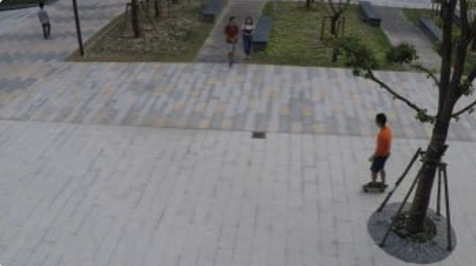}
\end{minipage}%
\hfill
\begin{minipage}{0.71\textwidth}
\textit{LLaVA:} ``The scene is set in a city park with a brick walkway, benches, and trees. However, the person on the skateboard is not wearing any pants, which is an unusual and unexpected sight.'' 

\textit{PandaGPT:} ``The couple should be the main focus of the image, 
and they should be actively engaged in skateboarding. Additionally, the surrounding
environment should be typical of a public space, with no unusual or out-of-place objects
or activities. If any of these conditions are not met, the image would be considered an anomaly.''
\end{minipage}
}

For the VLMs + Mistrial category, using OWLViT and LLaVA as visual perception modules yields usable results, though they are still 6 to 10\% lower than using CogVLM. However, the results with BLIP-2 and RAM are not usable due to serious hallucinations. For example, in a normal frame featuring only people walking, BLIP-2 outputs ``A sidewalk with trees, two people are walking down a sidewalk, a man is riding a skateboard on a sidewalk, a woman walking down a sidewalk in a park.'', while RAM (recognize anything) outputs ``Image Tags:  path | person | skate | park | pavement | plaza | skateboarder | walk''.

For the CogVLM + LLMs category, GPT-3.5 performs well but is expensive for large-scale testing. LLaMA-2, on the other hand, struggles with reasoning and fails to follow the given rules as context effectively.

In summary, the propose \sys is a generic plug-and-play framework that can improve VAD performance upon both the closed-source GPTs and the open-source VLMs/LLMs such as CogVLM and Mistral. \sys applies to various VLMs/LLMs backbones as long as they have decent visual perception and rule-following capabilities.

\subsection{Further Discussions on Perception Smoothing and Robust Reasoning}
Sections~\ref{sec:sota} and~\ref{sec:ab} demonstrate the effectiveness of the proposed perception smoothing and robust reasoning strategies. In this section, we provide a deeper investigation into them. Specifically, we aim to examine the extent to which the smoothing step may incorrectly smooth out anomalies from a sequence of video frames, and the extent to which the robust reasoning step can rectify these errors.

Table~\ref{tab:a-smoothing} shows that less than 0.7\% of anomalies are incorrectly smoothed out by the perception smoothing step (before the robust reasoning step), indicating very low false negative rates. The subsequent robust reasoning step successfully rechecks and corrects inaccuracies in the smoothed results, further reducing the false negative rates to below 0.15\%.

\begin{table}[ht]
\setlength{\tabcolsep}{17pt}
    \scriptsize
    \centering
    \caption{The percentage (\%) of incorrectly smoothed-out anomalies by the perception smoothing strategy on each dataset.}
    \label{tab:a-smoothing}
    \begin{tabular}{c|ccccc}
    \toprule
    Dataset  & ShT & Ave & Ped2 & UB \\
    \midrule
    Before Robust Reasoning & 0.7\% &0.4\% &0.6\% &0.3\%\\
    After Robust Reasoning & 0.08\% &0.15\% &0.08\% &0.01\%\\
    \bottomrule
    \end{tabular} 
\end{table}

The low false negative rates are due to that the smoothing step only smooths out the brief, isolated frames within a sequence of continuous frames. Table~\ref{tab:a-statistic} shows that brief anomalies are rare in VAD datasets, as they typically persist for 97.9 to 441.3 continuous frames due to the time required for an anomaly to enter and exit the camera's view. We also calculated the percentage of brief frames, i.e., $\leq 10$ frames, among all continuous anomaly frames. The ShT dataset has the highest percentage at 17.5\% and an average length of 5.5 frames. In Section~\ref{sec:ab}, we find that a padding size $p = 5$ in our majority vote step is the optimal window size for ShT for capturing the predominant motion continuity in a video. This aligns with the average length of brief continuous anomalies (5.5 frames) and may explain the reason behind this optimal value.

\begin{table}[ht]
\setlength{\tabcolsep}{12pt}
    \scriptsize
    \centering
    \caption{Statistics for the number of continuous anomaly frames per video clip of each dataset.}\label{tab:a-statistic}
    \begin{tabular}{c|ccccc}
    \toprule
    Dataset  & ShT & Ave & Ped2 & UB \\
    \midrule
    \# Average continuous anomaly frames & 111.3 & 97.9 & 137.3 & 441.3 \\
    \% Brief continuous anomalies ($\leq 10$ frames) & 17.5\% & 2.1\% & 0.0\% & 0.0\%  \\
    \# Average brief continuous anomaly frames & 5.5 & 10.0 & 0.0 & 0.0 \\
    \bottomrule
    \end{tabular} 
\end{table}

\subsection{Normal Reference Frame Sampling}
The proposed few-normal-shot prompting method is particularly beneficial when only a few normal data points are available in real-world scenarios. In our experiments, we simulate this scenario by randomly sampling normal reference frames from a training set, assuming only the randomly sampled frames are available.

However, even when a set of normal data (e.g., a training set) has already been collected, our few-normal-shot prompting method is still useful for fast adaptation. In this scenario, different normal reference frame sampling strategies beyond random sampling can be considered, such as sampling by GPT-4V~\cite{llm_achiam2023gpt4}. Table~\ref{tab:a-sampling} compares the random sampling and GPT-4V sampling (sampling ten frames) on the ShT dataset. The results of five trials show similar performance. The reason is that normal patterns in existing VAD datasets are not very diverse. Hence, randomly sampled normal frames are efficient as references for rule induction. Requiring only a few randomly sampled reference frames is one of our contributions, but GPT-4V sampling could be a promising extension for more complicated VAD scenarios.

\begin{table}[ht]
\setlength{\tabcolsep}{27pt}
    \scriptsize
    \centering
    \caption{Random sampling vs. GPT-4V sampling on the ShT dataset. Results of five trials are reported.} \label{tab:a-sampling}
    \begin{tabular}{c|c|c}
    \toprule
    Method & \# Rules & AUC (\%) \\
    \midrule
    Random sampling (ten frames) & 42.2 $\pm$ 4.2 & 84.5 $\pm$ 1.1 \\
    GPT-4V sampling (ten frames) & 39.9 $\pm$ 6.9 & 84.8 $\pm$ 1.6 \\
    \bottomrule
    \end{tabular} 
\end{table}

\newpage
\subsection{Unified Anomaly Detection}
Unified anomaly detection~\cite{you2022unified} considers image anomaly detection that trains a single model across different object classes. We extend this setting to VAD by considering a single model across different datasets. Specifically, the proposed \sys can perform as a unified anomaly detection approach by using normal reference frames randomly sampled from various datasets and deriving a set of unified rules for all datasets. Table~\ref{tab:uni} shows the results, which are on par with the main evaluation in Table~\ref{tab:sota}. This demonstrates that \sys performs well under the unified anomaly detection setting by inducing effective unified rules across datasets with similar anomaly scenarios but distinct visual appearances.

\begin{table}[ht]
\centering
\setlength{\tabcolsep}{35pt}
\scriptsize
\centering
\caption{AUC (\%) of \sys under the unified anomaly detection setting. \sys induces unified rules from a few normal reference frames across all four datasets and is evaluated on these datasets.} 
\label{tab:uni}
\begin{tabular}{cccc}
\toprule 
\textbf{Ped2} & \textbf{Ave} & \textbf{ShT} & \textbf{UB} \\
\midrule
97.6 & 85.6 & 84.7 & 68.8 \\
\bottomrule
\end{tabular}
\end{table}

\end{document}